# Data-Driven Predictive Modeling of Microfluidic Cancer Cell Separation Using a Deterministic Lateral Displacement Device

Elizabeth Chen
McKelvey School of Engineering
Washington University in St. Louis
St. Louis, MO, USA
chen.e.k@wustl.edu

Andrew Lee
School of EECS
Pennsylvania State University
University Park, PA, USA
ayl5530@psu.edu

Md Tanbir Sarowar
J. Mike Walker '66 Department of Mechanical Engineering
Texas A&M University
College Station, TX, USA
tanbirsarowar@tamu.edu

Xiaolin Chen
School of Engineering and Computer Science
Washington State University Vancouver
Vancouver, WA, USA
chenx@wsu.edu

*Abstract*— Deterministic Lateral Displacement (DLD) devices are widely used in microfluidics for label-free, size-based separation of particles and cells, with particular promise in isolating circulating tumor cells (CTCs) for early cancer diagnostics. This study focuses on the optimization of DLD design parameters—such as row shift fraction, post size, and gap distance—to enhance the selective isolation of lung cancer cells based on their physical properties. To overcome the challenges of rare CTC detection and reduce reliance on computationally intensive simulations, machine learning models including gradient boosting, k-nearest neighbors, random forest, and multilayer perceptron (MLP) regressors are employed. Trained on a large, numerically validated dataset, these models predict particle trajectories and identify optimal device configurations, enabling high-throughput and cost-effective DLD design. Beyond trajectory prediction, the models aid in isolating critical design variables, offering a systematic, data-driven framework for automated DLD optimization. This integrative approach advances the development of scalable and precise microfluidic systems for cancer diagnostics, contributing to the broader goals of early detection and personalized medicine.

## I. Introduction

Cancer remains a leading cause of death globally, necessitating continuous improvement in early diagnostic technologies. Traditional diagnosis on tissue biopsies relies on surgically removed tumor tissues. This approach has several limitations that hampers effective early diagnosis: the invasive nature causes patient discomfort and complication in process, and tumors must be in sufficient size for imaging detection. In contrast, diagnosis through liquid biopsies through detection of circulating tumor cells (CTCs), cell-free DNA (cfDNA), or other biomarkers within blood samples, enables effective early detection using a non-invasive technique [1].

Among emerging liquid biopsy techniques, circulating tumor cells–tumor-derived cells found in peripheral blood–have gained significant attention as biomarkers for early cancer detection, prognosis, and therapeutic monitoring. However, their rarity among a vast number of hematological cells poses substantial challenges for reliable and scalable isolation.

Deterministic Lateral Displacement (DLD) is a microfluidic technique that enables label-free (requiring no chemical modification or tagging), size-based separation of particles, and has been demonstrated to effectively isolate CTCs based on their distinct physical properties such as size and deformability [2], [3], [4]. The separation mechanism relies on the precise geometric design of micro-post arrays within the flow channel, with parameters such as row-shift fraction, post diameter, and gap spacing dictating the critical particle size [5]. Despite its advantages, the design and optimization of DLD devices remain limited by the need for expensive, time-consuming fabrication cycles and numerical simulations.

To address these constraints, recent efforts have turned to machine learning (ML) as a tool for accelerating device design and improving predictive accuracy. ML algorithms, trained on simulation or experimental data, can uncover complex relationships between design parameters and separation outcomes, reducing the need for exhaustive numerical simulations [6], [7]. Integrating ML into the DLD design workflow allows rapid iteration, high-throughput analysis, and the development of intelligent microfluidic platforms tailored to specific diagnostic requirements [7].

In this study, a hybrid modeling framework predicted cell trajectories and optimized design parameters by coupling high-fidelity numerical simulations of cell migration in DLD devices with supervised ML models. The study's aim is to use a data-driven approach to accelerate the design of DLD devices for effective CTC isolation, particularly for lung cancer diagnostics. Our results show that ML-based models not only approximate simulation outputs with high accuracy but also provide insight into critical parameter sensitivities, thus enabling more efficient and accessible device design.

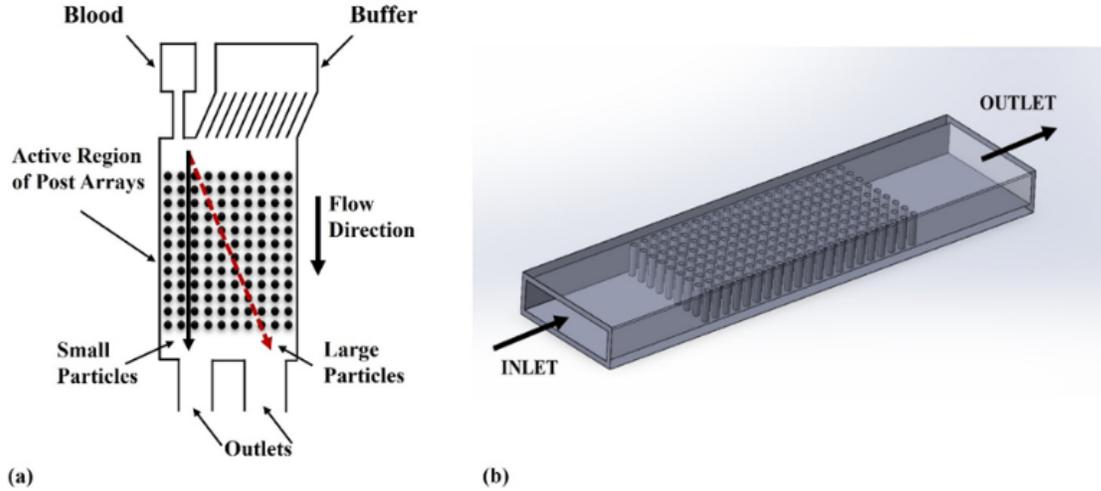

**Figure 1.** Schematics of a DLD device (a) Working Regions (b) 3D model generated in COMSOL

## II. DLD Theory

### A. Geometric Model

The DLD device separates particles of different sizes by placing an array of obstacles along the flow direction. The particles flowing through the fluid medium experience an asymmetry between the average fluid streamlines and the axis of placement of the obstacles. The configuration of this array placement causes a lateral shift in perpendicular to the average fluid flow for particles larger than a critical size. Figure 1(a) illustrates the main components of a DLD separation technique, namely an input region, an array region, and output region.

The input side consists of a fluidic channel to deliver the sample accompanied by the buffer flow. Generally, the buffer solutions comprise of different salts to support the particles/cells, as these are removed from the sample solution. Within the array region, the fluid flows between the obstacles from input to output, small cells follow the fluid streamline, and the larger cells deviate at an angle with respect to the fluid, as will be discussed in more detail later. The output region comprises of several segments that are separated to collect individual cells. Figure 1(b) represents the typical modeling approach for a DLD numerical study, that can comprise of an Inlet section, the array arrangement, and the outlet. The particles can be precisely released from any location of the inlet region with fluid solution and consequently their coordinate can be obtained at the outlet to study the lateral shift generated by the post's arrangement.

### B. Maintaining the Integrity of the Specifications

Initially introduced by Huang et al. [2] the DLD principle relies on periodic arrays of posts arranged to induce lateral shifts in fluid streamlines. Particles smaller than a critical diameter ($D_c$) follow fluid streamlines (zigzag mode), while larger particles shift laterally between streamlines (bumped mode) due to their size exceeding the width of fluid lanes (Figure 2).

The $D_c$, a key design parameter, distinguishes these two modes and depends primarily on the geometric arrangement of the device:

$$D_c = 2\alpha G \varepsilon \quad (1)$$

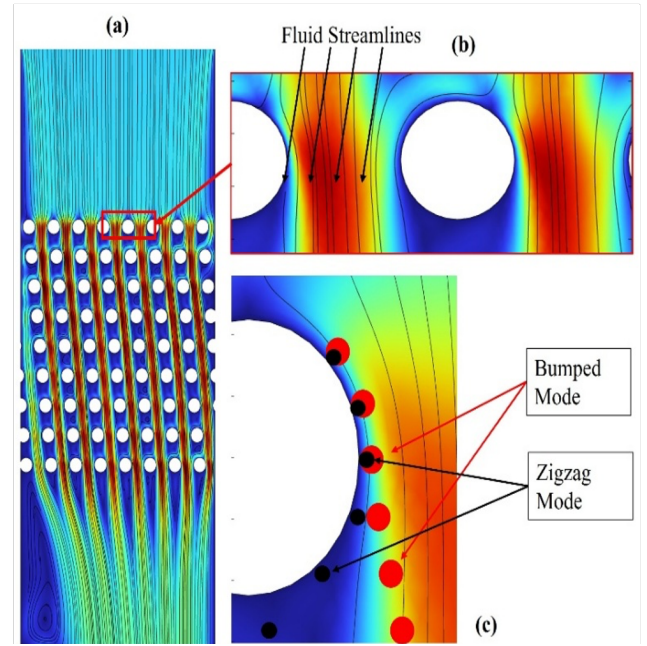

**Figure 2.** The orientation of fluid streamlines in a DLD device and its principle of operation. (a) the principle is presented with a lateral shift in flow streamlines produced in a device with 9 rows of obstacles. (b) the fluid streamline separation between two pillars is also presented. (c) the particle separation principle is shown with two different sizes of cells.

where $\alpha = \sqrt{(N/3)}$ accounts for non-uniform velocity profiles, G is the gap between posts. ε is the lateral shift fraction defined as:

$$\varepsilon = \frac{\Delta\lambda}{\lambda} = \frac{1}{N} \quad (2)$$

with $\lambda = G + D_p$ as the post center-to-center spacing (Figure 3) and N, the period indicating row repetition. Practically, Davis et al. [3] empirically refined this to provide improved predictions aligning closely with experimental outcomes:

$$D_c = 1.4G\varepsilon^{0.48} \quad (3)$$

C. *Bumped and Zigzag Modes*

As described in the preceding section, the particle transport in a DLD device follows two distinct modes: bumped and zigzag. As the name implies, in bumped mode, particles larger than $D_C$ shift laterally at each post interaction, moving at an angle (θ) relative to fluid flow direction, determined by the periodic lateral shift of micro post rows (Figure 3). Conversely, smaller particles remain confined within their initial streamline, undergoing a zigzag trajectory around posts without net lateral displacement.

D. *Physics of Flow Separations*

Fluid streamline separation within DLD devices arises due to the geometric arrangement of micro posts. The velocity profile between micro posts exhibits no-slip conditions at surfaces and maximum velocity at the channel center, creating unequal streamline widths. Larger streamlined widths near post surfaces accommodate greater fluid volumes to satisfy boundary conditions, while central streamlines are narrower and faster (Figure 2(b)). Particle trajectories depend critically on these streamline distributions.

E. *Incorporation of Machine Learning into Design*

DLD devices work on the principle of separating particles based on particle separation through placing obstacles in a tilted plane along the flow direction, exploiting the difference of forces acting on the particle depending on the size of the particles. This phenomenon, although following physics of continuum fluid mechanics, is often quite complex to model experimentally, requiring high speed and high-resolution cameras since the particles of interest are of micrometer dimensions. Consequently, during the initial design stages of a DLD device, numerical simulations are preferred over experimental modelling. Numerical simulations also impose complexities that are relevant in micro scale geometric domains, such as tracking particle trajectories at very small-time intervals (most commonly in the order of 10 microseconds). Also of importance, the design principles are highly dependent on parametric sweeps of crucial design parameters. This is not suitable for real-time design optimization and adjustments, but often required on clinical applications of cancer cell separations. The machine learning based predictive design approach offers an efficient alternative to commonly adopted DLD device design progressions. This includes predicting particle trajectory patterns in a hypothetical DLD design to optimize the design parameters.

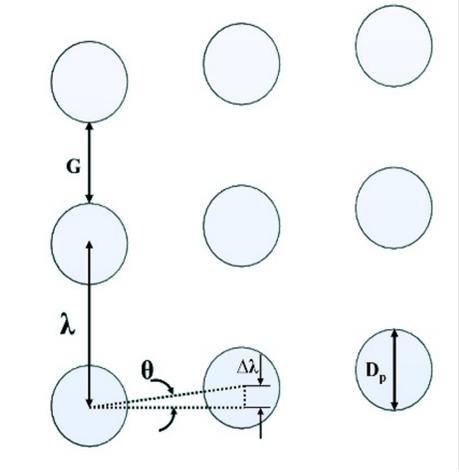

**Figure 3.** Row shifting in a DLD array design.

### III. METHODOLOGY

A. *Numerical Modeling*

Following the physics of particle transport through fluids traveling within a DLD device, a single particle's transport profile can be achieved through numerically solving the flow physics combining with the forces acting on the particle. The numerical solution can be regarded as the starting point of any DLD device design approach, which is useful to study particle transport profile at the initial stages. Thus, numerical solutions of particle transport for a large set of design parameters are generated in this study utilizing COMSOL Multiphysics 6.0. It is one of the most widely used numerical solvers for modeling flow physics [8]. It provides a robust and powerful platform to solve complex numerical problems combining multiple physics, for example, predicting a particle traveling through moving fluid through a microfluidic device. This kind of problem requires solving complex partial differential equations for fluid flow that accounts for Newton's laws of motion forces, acting on the particle at different time intervals.

The laminar flow module is used to solve the incompressible Navier-Stokes and continuity equations that govern the fluid field distribution across a DLD device:

$$\rho \left(\frac{\partial v}{\partial t} + v \cdot \nabla v\right) = \nabla \cdot (-pI + \mu(\nabla v + (\nabla v)^T)) + F_v \quad (4)$$

$$\nabla \cdot v = 0 \quad (5)$$

Cell motion was governed by forces including drag and lift, solved using Newton's second law:

$$m_{cell}\frac{d}{dt}(v_{cell}) = F_{Drag} + F_{Lift} \quad (6)$$

Drag force used the Schiller-Naumann model:

$$F_{Drag} = \left(\frac{1}{\tau_p}\right) m_{cell} (u - v_{cell}) \quad (7)$$

Lift force near walls was modeled by:

$$F_{Lift} = C_L \rho_f \frac{v_{cell}^2 d^2}{2} \quad (8)$$

Collision modeling enabled accompanying bouncing theory, including steric effects which were implemented by imaginary walls around obstacles. Constant velocity and pressure boundary conditions were set at inlet and outlet, respectively. A fully coupled iterative GMRES solver with generalized alpha automatic time stepping method was used. The computational mesh was optimized through grid independence tests. The hardware specifications included an Intel Xeon processor (Xeon E-2224) with 64 gigabytes of RAM.

*B. Data Generation for Machine Learning*

To train a machine learning model effectively, it is essential to generate sufficient data that capture the critical features of all design aspects, enabling higher accuracy in model predictions. One of the most important design aspects is the lateral displacement in the fluid streamlines, achieved by shifting each subsequent row laterally, referred to as the row shift. Therefore, data generation through numerical simulation focused on this parameter. As described in Section 2, the row shift fraction can be correlated with the inverse of the number of rows required for a complete period. This allows the design to be based on the number of rows per period, from which the lateral shift (row shift fraction) can be calculated. It also provides flexibility during the design stage, avoiding non-integer row numbers that would result in impractical designs.

TABLE I.  DESIGN PARAMETERS

| Parameters | Description | Value |
|---|---|---|
| $D_P$ | Post diameter | 45 μm |
| G | Gap distance | 45 μm |
| $R_N$ | Reynolds Number | 1 |
| N | Period Number | 3-48 |

The data generated based on the number of rows (period number, N) are summarized in Table 1. The post diameter and gap distance were both kept constant at 45μm and the flow Reynolds number was kept constant at 1. The number of rows varies from 3 to 48, as designs with fewer than 3 or more than 48 rows are unsuitable for DLD-based particle separation. Figure 4 (left) illustrates this process for N = 15, where 15 horizontal rows of obstacles define one period. Trajectories for three particle sizes are plotted, demonstrating that they provide the necessary information about particle transport modes. To fully capture the particle trajectory's transition from zigzag to bumped behavior, 28 different particle sizes ranging from 1-14μm were simulated for each DLD device design. Each of these simulations were then manually analyzed and labeled with the mode of particle transport through the DLD device according to the physics of particle movement: zigzag or bumped. Altogether, after removing simulations with inconclusive trajectory behavior, this process produced 1160 simulation datasets, each with approximately 10000 coordinate observations.

*C. Machine Learning*

The original simulation datasets were stratified split 80/20 into a training set and a testing set using the bumped or zigzag mode for stratification. This ensured equal representation of both modes across datasets and prevented data leakage between methodologies. Two different methodologies of applying ML to predict particle trajectory were evaluated: particle trajectory regression and trajectory mode (zigzag or bumped) classification. For particle trajectory regression, the study evaluated the following ML models: gradient boosting, k-nearest neighbors (kNN) and random forest. For particle trajectory mode classification, the study assessed kNN and multilayer perceptron (MLP) ML models.

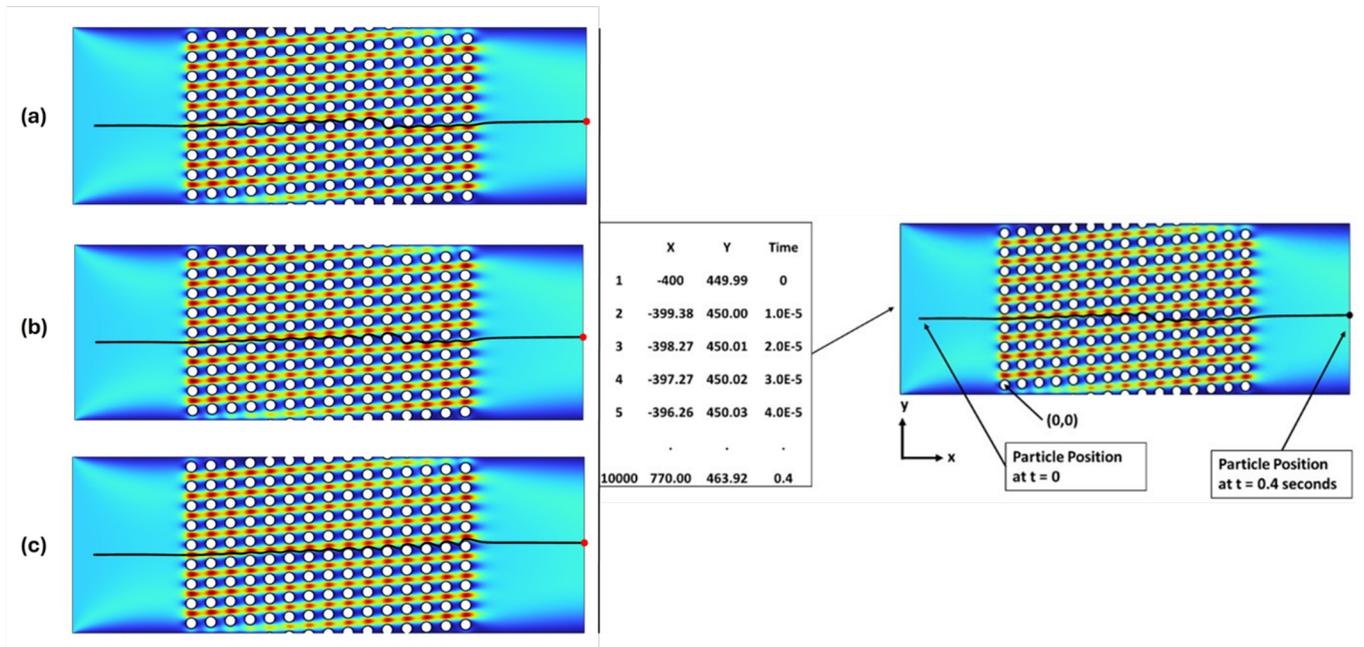

**Figure 4.** Particle trajectories obtained from numerical simulations for a DLD device with 15 period number (left). The example of how particle trajectory data has been stored as particle coordinates at different time-steps in the simulation domain.

All models employed hyperparameter tuning with stratified 5-fold cross validation to enhance prediction. Each model's predictive performance was rigorously evaluated using their respective evaluation metrics on the testing data to ensure reliability for practical DLD device applications. All models ran on an Apple M2 8-core CPU chip with 16 gigabytes of RAM.

*1) Particle Trajectory Regression*

For the particle trajectory regression task, each individual coordinate was labeled with the design parameter and particle size (Figure 4, right). ML models that applied three different methods—gradient boosting, kNN, and random forest—were trained to predict a particle's y-coordinates as it passed through a given DLD device design. Predictions used the particle's x-coordinate, particle size, and the design parameter of the DLD device. Gradient boosting models sequentially improve accuracy by minimizing residual errors; kNN predicts outputs based on proximity to neighboring data points; random forest leverages multiple decision trees for robust prediction. An alternative methodology of image processing was considered but not investigated due to time and resource constraints.

Regression model accuracy was evaluated using coefficient of determination ($R^2$), which is defined as follows:

$$R^2 = 1 - \frac{\sum_{i=1}^{n}(y_i - \hat{y}_i)^2}{\sum_{i=1}^{n}(y_i - \bar{y})^2} \quad (9)$$

where $y_i$ represents the COMSOL generated values, $\hat{y}_i$ represents the predicted value, $\bar{y}$ is the mean of the generated values, and $n$ is the number of data points. The higher $R^2$ with the maximum of 1 represents better model fit and accuracy.

*2) Trajectory Mode Classification*

For trajectory mode classification (i.e, to predict either bumped or zigzag mode), the study compared kNN and MLP models. MLP captures complex, non-linear relationships through neural network layers. As described earlier, kNN predicts outputs based on proximity to neighboring data points. Trajectory mode classification model accuracy was evaluated using precision, recall, F1-score, accuracy, and confusion matrix metrics, which are defined as follows:

$$Precision = \frac{True\ Positives}{True\ Positives + False\ Positives} \quad (10)$$

$$Recall = \frac{True\ Positives}{True\ Positives + False\ Negatives} \quad (11)$$

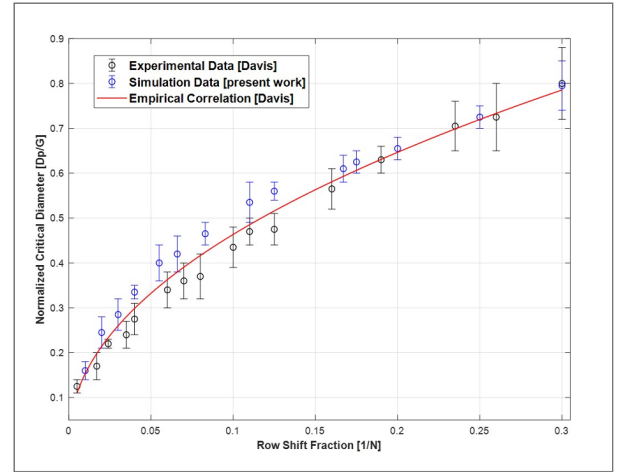

**Figure 5.** Comparison of critical diameter estimation by numerical simulations with experimental data and empirical correlations presented in Davis [3]

$$F1 - score = \frac{2 \times Precision \times Recall}{Precision + Recall} \quad (12)$$

$$Accuracy = \frac{True\ Positive + True\ Negative}{All\ Predictions} \quad (13)$$

IV. RESULTS

The results section presents the predictive modeling of a DLD device, including data generation via numerical simulations, data preparation for ML applications, and device design predictions using several ML models.

*A. Numerical Simulation*

The results from numerical simulations were validated against experimental data and the empirical formula by Davis, as shown in Figure 5. The critical diameter obtained from simulations was normalized by dividing by the gap distance between posts (G). Comparisons with the empirical correlation showed very good agreement. It is important to note that the critical diameter represents the threshold at which particle transport shifts from zigzag to bumped mode, making it difficult to pinpoint an exact particle size from simulations or experiments. Instead, a range is typically reported, determined by the smallest particle size increment used in the study. Additionally, this range (illustrated with error bars in Figure 5) is not uniform despite consistent particle size increments, due to the presence of a mixed transport mode, termed the "alternate zigzag" mode by Kim et al. [9]. The maximum difference between critical diameter estimates from numerical simulations and Davis's empirical formula was only 0.03 μm, corresponding to less than a 6% discrepancy.

*B. Particle Trajectory Prediction in a DLD Device*

Applying ML to predict particle trajectory in a DLD device using the particle regression model presents a promising method for streamlining the device design stage, requiring fewer simulations and experimental trials. Considering the complexity of the tasks, the regression models show comparatively excellent results, and the best performing model was the kNN regression model with an average $R^2$ training score of 0.979 and average $R^2$ testing score of 0.961. Figure 6 represents the particle trajectories predictions from the kNN model for a zigzag trajectory (left) and for a bumped trajectory (right).

Comparing results for kNN and MLP models that predicted zigzag or bumped classifications show that the MLP model was more successful in predicting particle transport mode. The MLP model uses artificial neural networks with multiple layers of neurons to learn patterns in data, making it well-suited for this type of classification task. The MLP model achieved training and testing accuracy of 97.4% and 98.7%, respectively. Figure 7 shows the confusion matrices for the training and testing datasets.

## V. Discussion

The results demonstrate the successful integration of numerical simulations with ML to create a robust predictive framework for DLD device design optimization. Validation of the numerical simulations against experimental data and empirical correlations (Figure 5) showed discrepancies below 6%, indicating the reliability of the simulations and supporting their use for training ML models.

### A. Model Performance

Among the evaluated ML models, the kNN regressor showed the best performance for particle trajectory predictions (Figure 6) with an average $R^2$ training score of 0.979 and average $R^2$ testing score of 0.961, while the MLP achieved the highest accuracy for classifying particle transport modes, with 97.4% training accuracy and 98.7% testing accuracy.

The success of kNN is attributed to its ability to capture complex, non-linear relationships between design parameters and particle behavior. Similarly, the MLP's strong performance in classification tasks reflects its capability to learn intricate patterns through multiple hidden layers, effectively distinguishing between zigzag and bumped modes based on device geometry and particle size.

### B. Particle Implications

The developed framework addresses critical challenges in designing DLD devices by significantly reducing the computation burden and time requirements associated with parametric studies. Traditional optimization requires extensive numerical simulations for each design iteration, making real-time clinical applications impractical. The ML solution enables rapid parameter screening and optimization through a pretrained framework, facilitating the development of personalized microfluidic devices tailored to specific cancer cell types and patient requirements.

### C. Limitations and Future Work

While this study's results are promising, several limitations arose that should be acknowledged. The current study focuses on a specific range of design parameters (N and particle sizes) with fixed flow conditions (Reynolds number = 1). Training ML models also requires extensive preparation of datasets generated from costly numerical simulations. Additionally, the regressors are trained on a full period of a DLD device, which increases the dataset size required for training. Future work could address these limitations by incorporating Navier-Stokes driven residuals to reduce the need for simulated datasets. In addition, predicting particle behavior within single unit of DLD device with flow fields would reduce the overall size of the dataset.

There is also much potential for exploring other methodologies such as image processing, applying transformer architectures for geometric modeling, or combining diffusion models with physic-informed generative methods to make trajectory predictions.

## VI. Conclusion

This study demonstrates the successful integration of numerical simulations with ML algorithms to establish a comprehensive predictive framework for DLD device optimization.

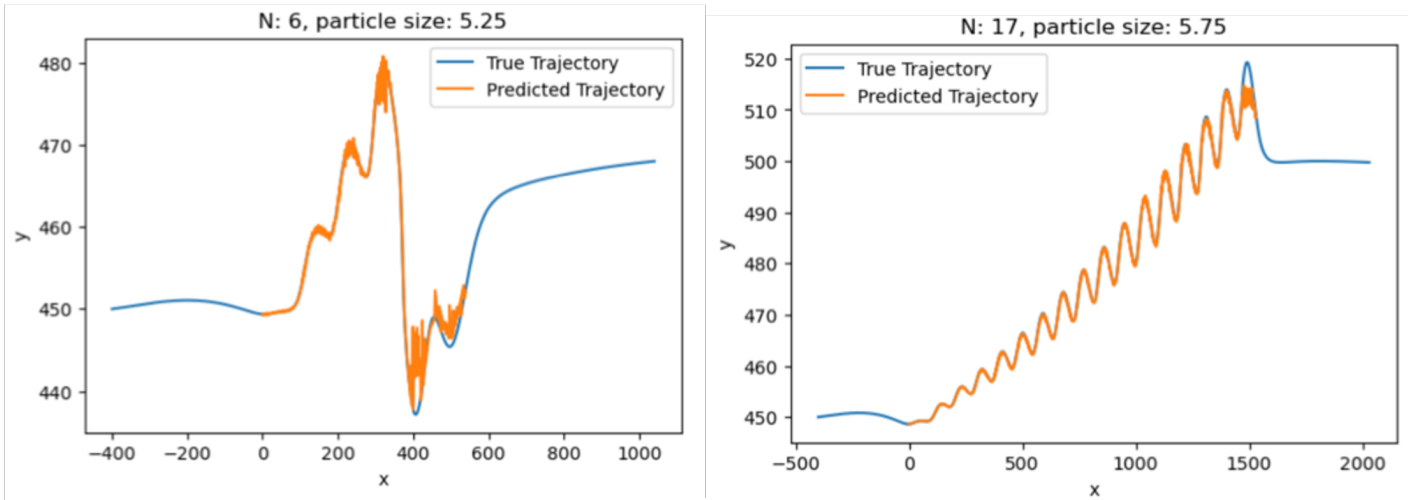

**Figure 6.** Comparison of particle trajectory predictions between kNN regression model with numerical simulations. (Left: performance with zigzag trajectory, right: performance with bumped trajectory)

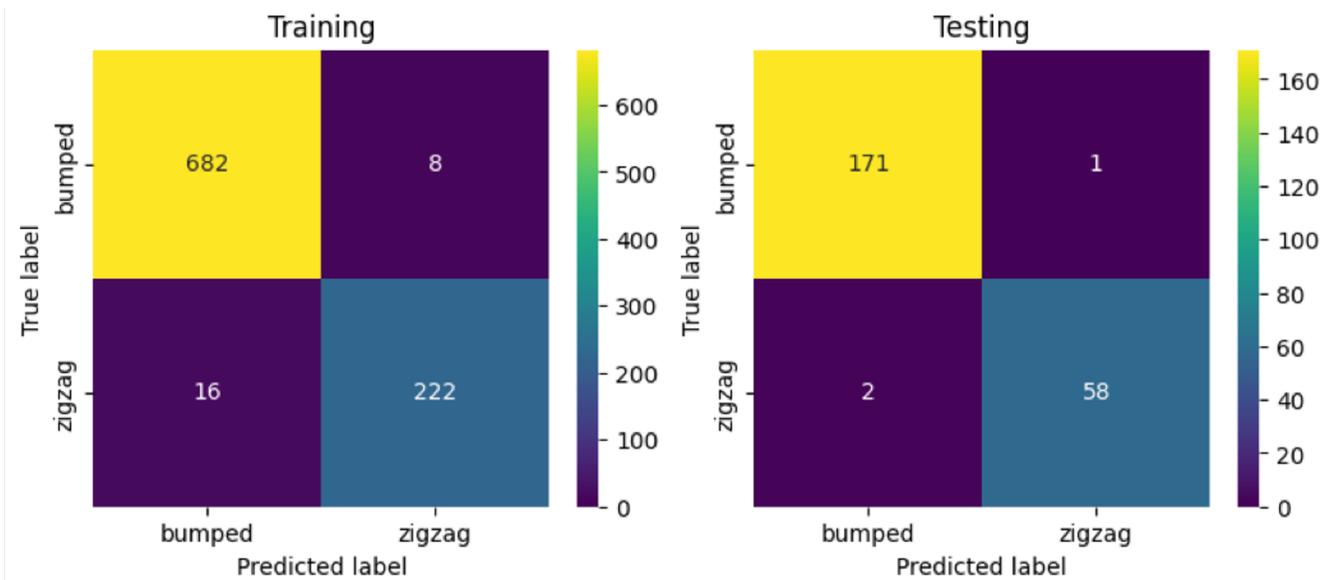

**Figure 7.** Supervised classification to predict mode of particle transport through DLD device with features (particle sizes and period number) and target (mode of particle transport).

The research confirmed that numerical simulations are stable and reliable for generating training datasets, with discrepancies below 6%.

The study identified two different ML methodologies to predict particle trajectory through a DLD device were evaluated: the particle trajectory regression method, and the particle trajectory mode (zigzag or bumped) classification method. For the particle trajectory regression method, a k-nearest neighbors (kNN) model excelled in predicting particle movement in an DLD device. For the particle trajectory mode classification method, a multilayer perceptron (MLP) model performed best in classifying zigzag or bumped trajectories. By capturing patterns through hidden layers, the ML approach enabled rapid parameter screening and optimization, significantly reducing the time and computational resources needed to iterate DLD device designs compared to traditional parametric studies. Future work should focus on minimizing dataset requirements by leveraging flow-field predictions and physics-driven residuals to optimize numerical simulation efforts. In conclusion, the proposed framework represents a notable advancement in microfluidic device development, offering a pathway toward improved early cancer diagnosis via liquid biopsies.

.